\documentclass[a4paper,fleqn]{cas-dc}

\usepackage[numbers]{natbib}
\usepackage{float}

\begin{document}
\let\WriteBookmarks\relax
\def\floatpagepagefraction{1}
\def\textpagefraction{.001}
\shorttitle{HP-GAN: Harnessing Pretrained Networks for GAN Improvement with FakeTwins and Discriminator Consistency}
\shortauthors{G. Son et~al.}

\title [mode = title]{HP-GAN: Harnessing Pretrained Networks for GAN Improvement with FakeTwins and Discriminator Consistency}                      

\author[1]{Geonhui Son}
\author[1]{Jeong Ryong Lee}
\author[1,2,3,4]{Dosik Hwang\corref{corres}}
\cortext[corres]{Corresponding author}

\affiliation[1]{School of Electrical and Electronic Engineering, Yonsei University, Seoul, Republic of Korea}
\affiliation[2]{Department of Radiology and Research Institute of Radiological Science and Center for Clinical Imaging Data Science, Yonsei University College of Medicine, Seoul, Republic of Korea.}
\affiliation[3]{Artificial Intelligence and Robotics Institute, Korea Institute of Science and Technology, Seoul, Republic of Korea}
\affiliation[4]{Department of Oral and Maxillofacial Radiology, Yonsei
University College of Dentistry, Seoul, Republic of Korea.}

\begin{abstract}
Generative Adversarial Networks (GANs) have made significant progress in enhancing the quality of image synthesis. Recent methods frequently leverage pretrained networks to calculate perceptual losses or utilize pretrained feature spaces. In this paper, we extend the capabilities of pretrained networks by incorporating innovative self-supervised learning techniques and enforcing consistency between discriminators during GAN training. Our proposed method, named HP-GAN, effectively exploits neural network priors through two primary strategies: FakeTwins and discriminator consistency. FakeTwins leverages pretrained networks as encoders to compute a self-supervised loss and applies this through the generated images to train the generator, thereby enabling the generation of more diverse and high quality images. Additionally, we introduce a consistency mechanism between discriminators that evaluate feature maps extracted from Convolutional Neural Network (CNN) and Vision Transformer (ViT) feature networks. Discriminator consistency promotes coherent learning among discriminators and enhances training robustness by aligning their assessments of image quality. Our extensive evaluation across seventeen datasets-including scenarios with large, small, and limited data, and covering a variety of image domains-demonstrates that HP-GAN consistently outperforms current state-of-the-art methods in terms of Fréchet Inception Distance (FID), achieving significant improvements in image diversity and quality. Code is available at: \url{https://github.com/higun2/HP-GAN}.
\end{abstract}




\maketitle

\section{Introduction}
\label{sec:intro}
Generative Adversarial Networks (GANs) \cite{goodfellow2020generative} are a prominent method in generative modeling, capable of synthesizing high quality, photorealistic images.
This capability is achieved through a min-max game between two components: a generator, which creates synthetic data, and a discriminator, which distinguishes between synthetic and real data.
Proper training of these two components is crucial for achieving high quality synthesis, as the discriminator acts as an adaptive loss function for the generator.

Nevertheless, GANs still encounter significant challenges such as non-convergence, training instability, and mode collapse \cite{arjovsky2017towards, mescheder2018training, zhang2019pa}.
These limitations are particularly pronounced when computational resources and datasets are limited, as is often the case in medical imaging of rare diseases, specific sets of celebrity portraits, or the artwork of a particular artist.
Previous studies have attempted to mitigate these issues through various approaches, including modifications in network architectures \cite{karras2019style, fastgan, radford2015unsupervised,sauer2021projected,zhang2019self}, refinement of objective functions \cite{arjovsky2017wasserstein, bellemare2017cramer, deshpande2018generative, kunkel2025wasserstein, li2017mmd, nowozin2016f}, regularization of weights and gradients \cite{arjovsky2017wasserstein, fedus2017many, mescheder2018training, miyato2018spectral, roth2017stabilizing, salimans2016improved}, and use of side information \cite{wang2018high, zhang2019variational, zhang2017stackgan, kim2021fat, kim2023dimix}.

Data augmentation, conventionally employed to mitigate overfitting in deep neural networks \cite{zhang2017mixup, cubuk2020randaugment, cubuk2018autoaugment}, has also been adapted to enhance GAN training.
Recent studies indicate that augmentation applied to both real and synthesized samples improves synthesis performance. \cite{karras2020training, zhang2019consistency, zhao2020differentiable, tran2021data, zhao2020image, zhang2019pa}.
However, the learning objective of the discriminator remains to distinguish between real and fake images, which degrade performance when training data is limited \cite{insgen}.

While transfer learning offers a potential solution \cite{mo2020freeze, wang2020minegan,ham2020unbalanced,wang2018transferring,zhao2020leveraging}, pretrained networks are not always compatible with the given training dataset, and improper fine-tuning can result in performance degradation \cite{zhao2020differentiable}.
Pretrained networks have extensive applications in image-to-image translation tasks, serving as perceptual loss functions \cite{johnson2016perceptual, dosovitskiy2016generating} or perceptual discriminators \cite{richter2022enhancing, sungatullina2018image}.
Projected GAN \cite{sauer2021projected} utilizes pretrained network to project generated and real samples into pretrained feature spaces, leading to significant improvements in performance, training stability, time, and data efficiency.
StyleGAN-XL \cite{stylegan-xl} utilizes both Convolutional Neural Networks (CNNs) and Vision Transformers (ViTs) networks for complementary effects.
Furthermore, Vision-aided GAN \cite{kumari2022ensembling} employs multiple networks by gradually increasing the number of pretrained networks during training.

In this paper, we propose HP-GAN, a novel approach that leverages pretrained networks to improve the performance of the generator by incorporating self-supervised learning (SSL) techniques.
First, we introduce FakeTwins, which utilizes pretrained feature networks as SSL encoders and trains the generator with SSL based on Barlow Twins \cite{barlow-twins} through the generated images.
Recently, contrastive learning based SSL methods, such as SimCLR \cite{chen2020simple} and MoCov2 \cite{chen2020improved}, have been integrated in GANs \cite{contraD, insgen, fakeclr}.
These methods incorporate contrastive learning as an auxiliary task during GAN training, thereby enhancing the discriminative capabilities of the discriminator.
In contrast, we leverage information maximization based methods \cite{barlow-twins, ermolov2021whitening, bardes2021vicreg} to enhance the capabilities of the generator in different way.
Barlow Twins is an SSL method that minimizes redundancy between the output features of two identical networks processing differently distorted (or augmented) versions of the same sample.
We assume that Barlow Twins with a pretrained network yields lower loss values for batches with diverse images, motivating the development of FakeTwins to enhance the diversity and quality of generated images.

Furthermore, we introduce the discriminator consistency loss, a regularization technique designed to ensure cohesive learning among multiple discriminators.
We employ both Convolutional Neural Network (CNN) and  Vision Transformer (ViT) as feature networks \cite{stylegan-xl}. 
However, due to the inherent architectural differences between CNNs and ViTs, the discriminator outputs from these networks do not align identically.
The discriminator consistency loss mitigates these discrepancies by encouraging a high level of consensus among the discriminators regarding the quality of images.
This approach ensures that the generator receives uniform and coherent feedback, leading to more stable and robust training.
In other words, the alignment of discriminator outputs leverages the structural differences between the pretrained CNN and ViT to provide a comprehensive assessment of the generated image.

To assess the quality of images produced by generative models, we employ several metrics for comprehensive comparison. Fréchet Inception Distance (FID) \cite{heusel2017gans} measures the disparity in distribution density within the feature space, providing a quantitative evaluation of image quality. Additionally, Kernel Inception Distance (KID) \cite{binkowski2018demystifying} is an unbiased alternative to FID, often yielding more stable evaluations for smaller sample sizes.
Precision and recall \cite{kynkaanniemi2019improved} offer additional insight by quantifying the proportion of generated images that resemble the training set and the model's ability to replicate the training set, respectively.

The main contributions of our work are summarized as follows:
\begin{enumerate}
\item We propose the discriminator consistency loss, a novel regularization technique that mitigates disparities in discriminator outputs.
Discriminator consistency improves training stability and overall performance by providing coherent feedback to the generator, ensuring robust and stable training.
\item We introduce FakeTwins, an innovative approach that incorporates self-supervised learning based method to train the generator. 
By utilizing pretrained feature networks as SSL encoders, FakeTwins enhances the diversity and fidelity of the generated images.
\item Through extensive experimental evaluations on seventeen diverse image-domain datasets, HP-GAN consistently outperforms existing state-of-the-art methods in generative modeling. Our method demonstrates significant improvements in image synthesis quality, achieving state-of-the-art performance across various benchmarks.
\end{enumerate}


\section{Related Works}
\subsection{Generative Adversarial Networks}
Generative Adversarial Networks (GANs) \cite{goodfellow2020generative} model the data distribution of a given training dataset. This is accomplished by setting up a min-max game involving two neural networks known as the generator and the discriminator.
The generator $G$ maps latent code $\mathbf{z}$ sampled from the latent space $\mathcal{Z}$, and produces realistic-looking samples $G(\mathbf{z})$.
The discriminator $D$ is fed either real samples $\mathbf{x}$  or synthetic samples $G(\mathbf{z})$ from the generator and tries to accurately distinguish them as real or fake. 
The min-max objective function of GAN is as follows:
\begin{equation}
\min _G \max _D\left(\mathbb{E}_{\mathbf{x}}[\log D(\mathbf{x})]+\mathbb{E}_{\mathbf{z}}[\log (1-D(G(\mathbf{z})))]\right).
\end{equation}

\subsection{Data Augmentation in GANs}
Data augmentation maximizes the utilization of datasets to improve performance on tasks such as classification \cite{zhang2017mixup, cubuk2020randaugment, cubuk2018autoaugment} by artificially increasing the size and diversity of training datasets.
In GANs, data augmentation also has been employed to prevent overfitting, stabilize training, and improve the quality of synthesized images, especially when a limited number of training samples are available \cite{karras2020training, zhang2019consistency, zhao2020differentiable, tran2021data, zhao2020image, zhang2019pa}. 

For instance, CR-GAN \cite{zhang2019consistency} augments the real images and introduces a consistency loss for training the discriminator. 
The other methods, ADA \cite{karras2020training} and DA \cite{zhao2020differentiable}, propose a concept of applying differential augmentation on both real and synthetic images to address the insufficiency of the training data and thus enhance the quality and diversity of the generated data.
ADA \cite{karras2020training} utilizes data augmentation through an adaptive technique. The strength of the augmentation is adjusted according to the current level of training stability, thus maintaining a balance between training stability and the diversity of the generated images. 
DA \cite{zhao2020image} introduces a differentiable augmentation strategy. This allows the gradients from the discriminator to flow back to the generator through the augmentations.

In a different direction, DANI \cite{zhang2024improving} introduces feature-space augmentation by injecting learnable noise into both the generator and discriminator. 
Instead of transforming input images, DANI simulates the regularization effect of augmentation through adaptive noise, improving stability and performance in low-data regimes.


\subsection{Self-supervised Learning}
Self-supervised learning (SSL) is an emerging paradigm that aims to learn meaningful representations from unlabeled data. SSL has demonstrated significant success in various domains, including image classification, object detection, and natural language processing \cite{chen2020simple, chen2020improved, he2020momentum, caron2020unsupervised, grill2020bootstrap, chen2021exploring, barlow-twins, bardes2021vicreg, gidaris2021obow, bachman2019learning, misra2020self, tian2020makes}
SSL methods typically learn representations that remain invariant under various transformations or different distortions.

\begin{figure*}[htbp]
  \centering
    \includegraphics[width=\textwidth, keepaspectratio]{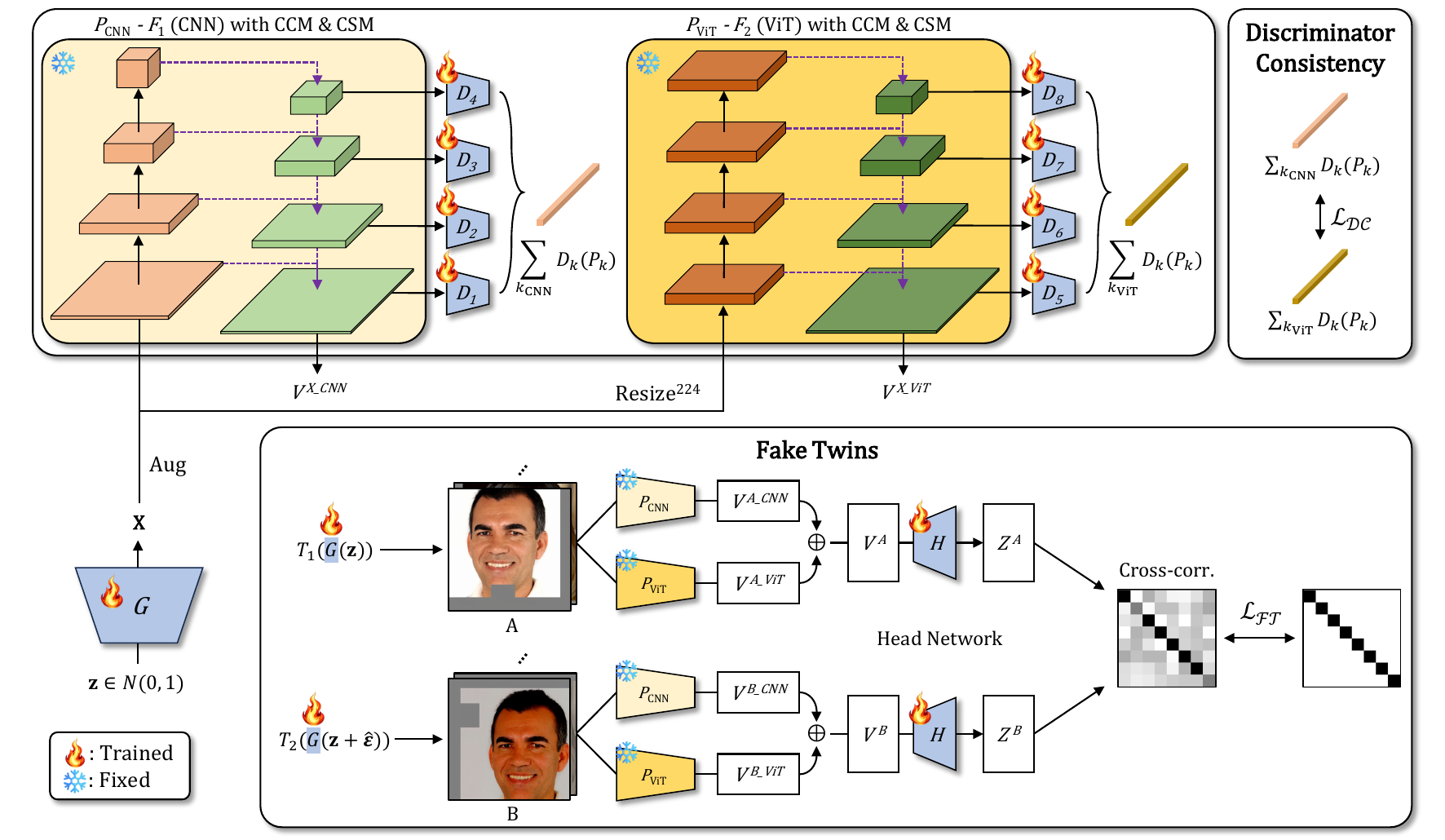}  
\caption{Algorithmic overview. The images are augmented and resized, then passed through pretrained CNN and ViT networks, along with their respective feature mixing blocks (CCM+CSM, dashed purple arrows). 
The processed multi-scale feature maps are then subjected to independent discriminators.
During training, we impose consistency between discriminator outputs from different feature networks ($\mathcal{L}_{\mathcal{DC}}$). Lastly, we propose the FakeTwins loss ($\mathcal{L}_{\mathcal{FT}}$), which is applied through fake images, with the aim of maximizing the diversity of images. FakeTwins is an SSL based approach which utilizes pretrained networks as encoders. $\oplus$ indicates concatenation.}
\label{main_fig}
\end{figure*}

In contrastive learning \cite{chen2020simple, chen2020improved, he2020momentum, misra2020self, InfoNCE}, the model learns an association between a positive pair (two augmented versions of the same image) and distinguishes it from a set of negative samples (augmented versions of different images). 
Contrastive methods have shown effectiveness, they also have limitations, such as the difficulty of selecting negative samples and the requirement of large batch sizes.

Information maximization methods \cite{barlow-twins, ermolov2021whitening, bardes2021vicreg, lee2025fedi} aim to maximize the information content of the embedding. These methods decorrelate each pair of embedding variables that contain redundant information, leading to the redundancy reduction that enhances the information content of the embedding vectors. Moreover, information maximization methods do not require negative samples, asymmetric networks, and large batch sizes.

Barlow Twins~\cite{barlow-twins} forces the normalized cross-correlation matrix of the two embeddings towards the identity.
More specifically, the procedure starts by generating two distorted views for images of a batch $X$ sampled from a dataset. Two batches of views $A$ and $B$ are created by different data augmentations ($T_1$ and $T_2$), and then encoded into representations $V^A$ and $V^B$ by the encoder network $f$. The representations are fed to the linear head network $\phi$, which produces the embeddings $Z^A$ and $Z^B$. 
The loss function is constructed to encourage decorrelation between the feature representations of the two augmented views, and mathematically represented as:
\begin{align}
&A=T_{1}(X), \quad V^{A}=f(A), \quad Z^{A}=\phi(V^{A}), \\
&B=T_{2}(X), \quad V^{B}=f(B), \quad Z^{B}=\phi(V^{B}),\\
&\mathcal{L}_{\mathcal{BT}}(Z^A, \, Z^B) \triangleq \sum_i{\left(1-\mathcal{C}_{i i}\right)}^2+\lambda_{1} \sum_i \sum_{j \neq i} \mathcal{C}_{i j}^2,
\label{Xeqn2}
\end{align}
where $\lambda_{1}$ is a constant that balances the importance of the first and second terms of the loss, and is set to $0.005$. $\mathcal{C}$ is the cross-correlation matrix between the outputs of the two identical networks, computed across the batch dimension:
\begin{equation}
    \mathcal{C}_{i j} \triangleq \frac{\sum_b z_{b, i}^A z_{b, j}^B}{\sqrt{\sum_b\left(z_{b, i}^A\right)^2} \sqrt{\sum_b\left(z_{b, j}^B\right)^2}},
\end{equation}
where $b$ indexes batch samples and $i, j$ index the vector dimension of the network’s output. $\mathcal{C}$ is a square matrix with the dimensions of the network’s output, and its values range between -1 (perfect anti-correlation) and 1 (perfect correlation).
As a result, the embedding vectors of distorted versions of a sample become similar, while simultaneously reducing the redundancy within vector components.

\begin{table*}[t]
\centering
\caption{Ablation study on FFHQ and Pokemon dataset. Results for different methods and configurations. Perceptual path length (PPL) \cite{karras2019style} is computed in $\mathcal{Z}$ only when $\mathbf{z} \in \mathbb{R}^{64}$, without the central crop. The best results are in bold.}
\label{config_table}
\resizebox{\textwidth}{!}{
\begin{tabular}{ll|cccc|cccc}
\hline
\multicolumn{2}{l|}{\multirow{2}{*}{Configuration}} & \multicolumn{4}{c|}{FFHQ (70k)}                                           & \multicolumn{4}{c}{Pokemon (833)}                                         \\
\multicolumn{2}{c|}{}                               & FID ↓         & PPL (full / end) ↓      & Pr ↑    & Rec ↑       & FID ↓          & PPL (full / end) ↓     & Pr ↑    & Rec ↑       \\ \hline
\textbf{A}   & FastGAN \cite{fastgan}                             & 12.69         & -                       & \textbf{0.716} & 0.184          & 81.86          & -                      & 0.731          & 0.004          \\
\textbf{B}   & + Projected GAN (CNN as $F_1$) \cite{sauer2021projected}          & 3.39          & -                       & 0.654          & 0.464          & 26.36          & -                      & \textbf{0.809} & 0.259          \\
\textbf{C}   & + ViT as $F_2$ \& small \textbf{z} \& blur reg. \cite{stylegan-xl}  & 2.29         & 150.3 / 150.1           & 0.672          & 0.497          & 24.70          & 568.5 / 574.3          & 0.793          & 0.164          \\
\textbf{D}   & + Discriminator consistency          & 1.86 & 134.3 / 133.4           & 0.664          & 0.524          & 24.02          & 548.9 / 554.8          & 0.788          & 0.210          \\
\textbf{E}   & + FakeTwins (HP-GAN)               & \textbf{1.69} & \textbf{132.50 / 131.9} & 0.642          & \textbf{0.545}          & \textbf{23.62} & \textbf{450.6 / 451.3} & 0.768          & \textbf{0.310} \\ \hline
\end{tabular}
}
\end{table*}

\subsection{Self-supervised Learning in GANs}
The incorporation of SSL in GANs is a recent area of research.
Tran et al. \cite{tran2019self} propose a multi-class minimax game where a rotation-based classifier distinguishes between transformed real and fake images, encouraging the generator to align with the self-supervised task. Hou et al. \cite{hou2021self} introduce label augmentation by integrating transformation labels into the adversarial objective, unifying self-supervised and GAN training through a single multi-class classification framework.

Recent approaches involve applying contrastive learning to GANs \cite{contraD, insgen, fakeclr, gou2024few, wang2024contrastive} and demonstrate that SSL loss improves the generation performance.
ContraD \cite{contraD} reformulates the conventional task to distinguish real and fake samples with two different contrastive losses: the SimCLR loss \cite{chen2020simple} on the real samples and the supervised contrastive loss \cite{khosla2020supervised} on the fake samples. 
InsGen \cite{insgen} introduces a self-supervised auxiliary task for the discriminator, aiming to enhance its discriminative power by following MoCo-v2 \cite{chen2020improved}. By distinguishing images via contrastive learning, synthesis performance is enhanced.
FakeCLR \cite{fakeclr} improves InsGen by introducing new strategies such as applying contrastive loss only to the fake images and implementing noise-related latent augmentation.
The implementation of contrastive loss also indirectly changes the generator's training objectives.
As a result, the discriminator and generator are trained better with information from instance discrimination, resulting in improved synthesis quality and diversity.

And for the conditional image generation, IG-GAN \cite{casanova2021instance} trains GANs conditioned on embeddings on SwAV \cite{caron2020unsupervised}.
ReACGAN \cite{kang2021rebooting} and ContraGAN \cite{kang2020contragan} introduce contrastive loss into discriminator training, enabling the discriminator to compare visual differences between multiple images and learn the fine-grained representation.
NoisyTwins \cite{rangwani2023noisytwins} introduces a self-supervised decorrelation loss inspired by Barlow Twins, applied in the latent space of class embeddings. However, this method inherently relies on the class labels and operates only in conditional settings. In contrast, our method applies the loss directly to image features extracted from high-capacity pretrained networks, enabling class-agnostic learning. This allows richer, perceptually-aligned redundancy reduction that guides the generator more effectively across a broader range of scenarios.

Unlike previous methods \cite{insgen, fakeclr, contraD, kang2021rebooting, kang2020contragan, casanova2021instance}, which typically treat the discriminator’s feature extractor as the trainable SSL encoder and mainly use contrastive, instance-discrimination losses, HP-GAN uses fixed pretrained networks as encoders and applies a negative-free, information-maximization SSL objective directly to generator outputs. Thus, FakeTwins directly regularizes the generator in a pretrained feature space, rather than only improving the discriminator’s discriminative power.

\subsection{Pretrained Models in GANs}
Pretrained models, especially those trained with large-scale supervised, unsupervised or self-supervised learning \cite{caron2020unsupervised,chen2020simple,he2016deep,krizhevsky2017imagenet,radford2021learning}, have shown impressive performance in a wide range of tasks through the use of useful representations, transfer learning, and fine-tuning \cite{donahue2014decaf,huh2016makes, kornblith2019better, oquab2014learning, saenko2010adapting, yosinski2014transferable, zeiler2014visualizing}.
Pretrained models have proven effective in reducing the need for extensive training datasets and training time.

Recent works have proposed the idea of leveraging pretrained representations for GANs, thus alleviating the burden of learning from scratch and accelerating convergence speed \cite{mo2020freeze, wang2020minegan, ham2020unbalanced, wang2018transferring, zhao2020leveraging}.
Another methods introduce perceptual loss function \cite{johnson2016perceptual, dosovitskiy2016generating}, which is used for conditional image synthesis tasks \cite{ledig2017photo, chen2017photographic, park2019semantic, wang2018high, gatys2016image}.
By comparing the feature representations of the real and generated images in the pretrained model's feature space, they guide the generator to generate images that are perceptually close to the real images. More recent methods such as DreamSim~\cite{fu2023dreamsim} learn human-aligned perceptual
similarity metrics from synthetic data and large pretrained feature spaces, providing stronger
feature-based distances for analyzing GAN representations.
Image-to-image translation is performed by combining perceptual loss and adversarial loss.
In addition, the perceptual discriminator \cite{richter2022enhancing, sungatullina2018image} that utilizes a pretrained VGG \cite{simonyan2014very} is proposed. 

On the other hand, pretrained models are used in different ways for unconditional GAN.
Projected GAN \cite{sauer2021projected} proposes the concept of projecting generated and real samples into a pretrained feature space.
The Projected GAN objective is formulated as follows:
\begin{equation}
\begin{aligned}
\min _{{G}} \max _{{D}_k} \sum_{k} & \Big(\mathbb{E}_{{\mathbf{x}}}\left[\log {D}_k\left({P}_k({\mathbf{x}})\right)\right]\Big. \\
& \Big.+\mathbb{E}_{{\mathbf{z}}}\left[\log \left(1-{D}_k\left({P}_k({G}({\mathbf{z}}))\right)\right)\right] \Big),
\end{aligned}
\end{equation}
where $P_k$ is a set of feature projectors and $D_k$ is a set of independent discriminators operating on different feature projections.
The feature projectors comprise a pretrained feature network $F$, along with cross-channel mixing (CCM) and cross-scale mixing (CSM) layers.
The aim of CCM and CSM is to dilute prominent features, preventing the discriminators from focusing on a certain subset of its input feature space. Both layers utilize differentiable random projections that are randomly initialized and not trained during GAN training. 
CCM mixes features across channels using random 1$\times$1 convolutions, and CSM mixes features across scales using residual random 3$\times$3 convolutions and bilinear upsampling.
The output of CSM is a feature pyramid with four feature maps at different resolutions, and four discriminators operate independently on these feature maps. 
StyleGAN-XL \cite{stylegan-xl} introduces a progressive growing strategy \cite{karras2017progressive} to StyleGAN3 \cite{karras2021alias}, demonstrating that incorporating multiple pretrained feature extractors improves generation quality.
Vision-aided GAN \cite{kumari2022ensembling} also validates the effectiveness of ensembling multiple pretrained networks.

However, these methods either use a single feature backbone or treat multiple
backbones independently, with each discriminator separately distinguishing
between real and fake images.
In contrast, our method goes further by leveraging pretrained networks not only for real/fake discrimination but also as encoders for self-supervised learning. This dual role of pretrained networks provides richer supervisory signals and promotes the generation of more diverse and high-quality images. Moreover, our method introduces an explicit consistency regularization term that aligns the outputs of heterogeneous discriminators, encouraging stable learning dynamics and coherent generator updates, which is conceptually related to consistency learning~\cite{guo2025multi}.


\begin{figure}[tbp]
  \centering
    \includegraphics[width=1\linewidth, keepaspectratio]{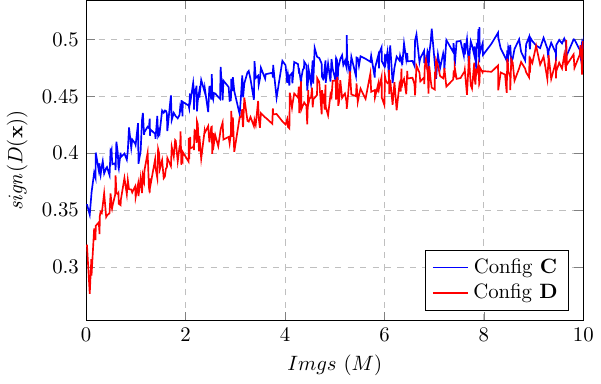}  
\caption{The signed real logits of the discriminator $sign(D(\textbf{x}))$ according to Config \textbf{D} and \textbf{E} on the FFHQ dataset. Imgs (M) denotes the cumulative number of training images (in millions).}
\label{sign_graph}
\end{figure}

\section{Methodology}
\label{Methodology}
In this section, we present our novel approach, which combines the capabilities of pretrained models with self-supervised learning (SSL) to produce diverse and high-quality synthetic images.
Our proposed method, termed Harnessing Pretrained Networks GAN (HP-GAN), builds on the FastGAN baseline (Config \textbf{A}), incorporating a series of progressive enhancements to improve performance and robustness.
The step-by-step introduction of these modifications is detailed below, with their respective impacts summarized in Table \ref{config_table}.

\begin{figure}[tbp]
  \centering
    \includegraphics[width=1\linewidth, keepaspectratio]{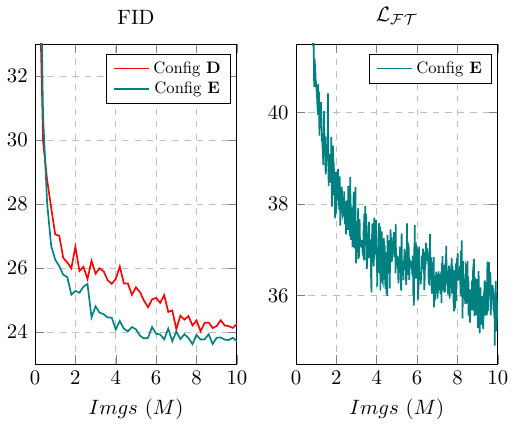}  
\caption{FID and $\mathcal{L}_{\mathcal{FT}}$ according to Config \textbf{D} and \textbf{E} on the pokemon dataset. Imgs (M) denotes the cumulative number of training images (in millions).}
\label{fid_graph}
\end{figure}

First, we restructure the traditional single discriminator setup into a multi-discriminator architecture, like Projected GAN (Config \textbf{B}). 
Next, taking cues from StyleGAN-XL, we introduce a hybrid architecture that integrates CNNs and ViT as feature networks (Config \textbf{C}). 

The subsequent enhancement involves the application of discriminator consistency (Config \textbf{D}). By ensuring that the outputs of discriminators, which are based on different feature networks (CNN and ViT), are aligned, we promote a higher level of consensus on image quality, leading to more uniform feedback to the generator and stable training.
Finally, we incorporate FakeTwins, resulting in Config \textbf{E}. By leveraging pretrained networks, we enhance the generator's ability to produce diverse images.
Our integrated framework, as depicted in Fig. \ref{main_fig}, demonstrates how these incremental innovations collectively contribute to significant improvements in GAN performance.

\subsection{Training GANs}
Our proposed method, HP-GAN, utilizes feature projectors with multiple feature networks and multi-scale discriminators.
We employ a hinge loss \cite{lim2017geometric} to train our $D$ and $G$. Concretely, loss functions are formalized as:

\begin{align}
\begin{split}
\mathcal{L}_D= \sum_{k}  & \Big(\mathbb{E}_{\mathbf{x}}[\text{max}(0, 1 - D_k({P}_k(\mathbf{x})))] \\ & + \mathbb{E}_{\mathbf{z}}[\text{max}(0, 1+D_k({P}_k(G(\mathbf{z}))))] \Big), 
\end{split}
\\ \mathcal{L}_G=\sum_{k} &-\mathbb{E}_{\mathbf{z} }[(D_k({P}_k(G(\mathbf{z}))))]. 
\label{eq:label} 
\end{align}

During the training process, we apply image blurring through a Gaussian filter with $\sigma$ = 2 pixels for the initial 200k images. 
Such discriminator blurring \cite{karras2021alias, stylegan-xl} mitigates the discriminator from focusing on high-frequency details in the early stages of training. 
In addition, following \cite{sauer2021projected}, we use spectral normalization without gradient penalties.

While conventional GAN approaches operate in high-dimensional latent spaces, recent studies suggest that natural image datasets possess a relatively lower intrinsic dimensionality \cite{pope2021intrinsic}.
StyleGAN-XL \cite{stylegan-xl} uses a low dimensional latent space and works more optimally.
With these insights, we reduce the dimension of the latent code $\mathbf{z}$ to 64, resulting in lower FID (Config \textbf{C}).

\begin{figure*}[tbp]
\centering
\includegraphics[width=1\textwidth]{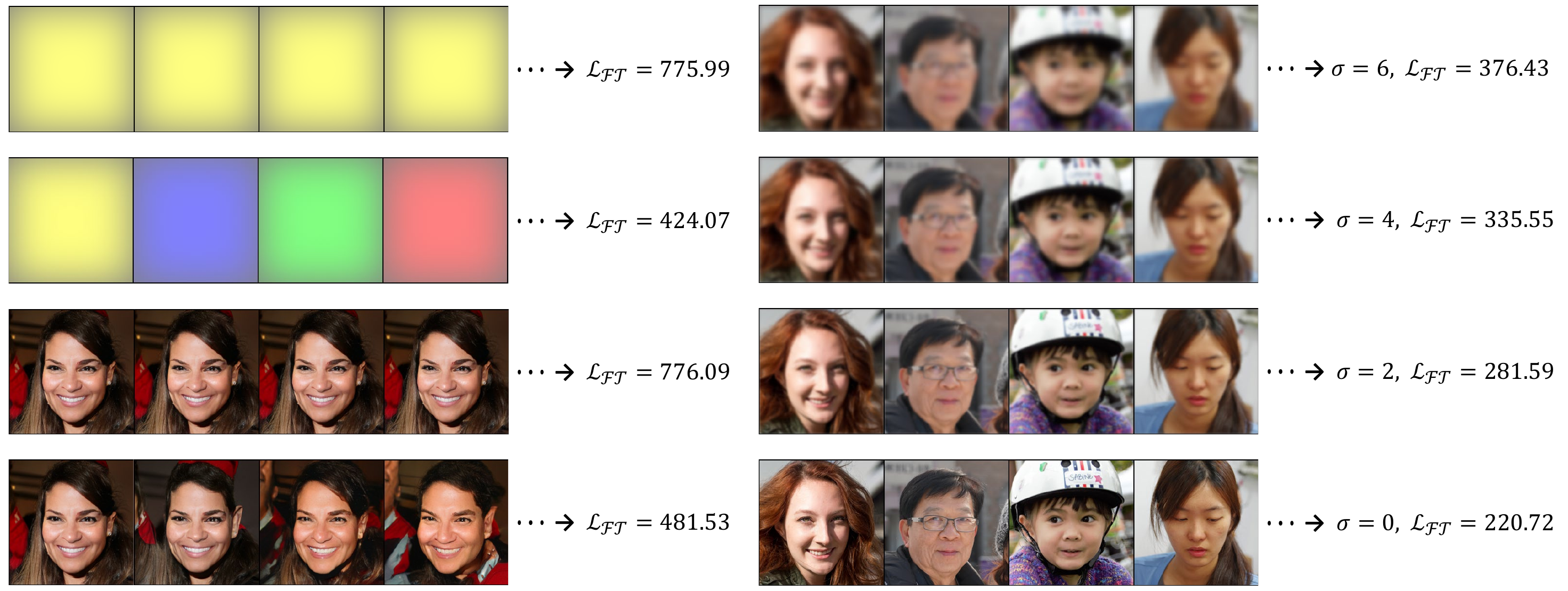}
\caption{$\mathcal{L}_{\mathcal{FT}}$ in batches of images with various settings. 
Rows 1 and 2 in the left column are the results when single color image is repeated, and when multiple color images are in the batch.
Rows 3 and 4 in the left column are when single face image is repeated, and when the batch consists of similar but different images generated with noise perturbation.
The right column is the comparison according to the degree of Gaussian blurring.
We compute 100 times with different random augmentation and report the average for each result.}
\label{batch_ft}
\end{figure*}

\subsection{Discriminator Consistency}
\label{D_Consistency}
In HP-GAN, discriminator consistency is introduced as a regularization technique to minimize the difference in discriminator outputs.  
The purpose of this consistency loss is to align the outputs of discriminators operating on feature maps extracted from different networks.
The discriminator consistency loss is formulated as follows:
\begin{equation}
    \mathcal{L}_{\mathcal{DC}}(\mathbf{x})=\mathbb{E}_{\textbf{x}} \bigg[\Big(\sum_{k_{\text{CNN}}}D_k({P}_k(\mathbf{x})) - \sum_{k_{\text{ViT}}}D_k({P}_k(\mathbf{x}))\Big)^2 \bigg]
\end{equation}
where $k_\text{CNN}$ and $k_\text{ViT}$ denote the set of $k$ discriminators within the CNN and ViT feature networks, respectively, as illustrated in Fig. \ref{main_fig}. During training, this loss is applied on both real images $\textbf{x}$ and generated images $G(\textbf{z})$.

Discriminator consistency leverages the distinctive image processing capabilities of CNNs and ViTs.
Due to the structure of the networks and the inherent differences in how each network processes the image between CNN and ViT, disparities in discriminator outputs are inevitable.
However, by aligning these outputs, we can achieve mutual agreement between discriminators regarding the quality of generated images.
Rather than enforcing feature alignment, which may suppress architectural diversity, we regularize the outputs to retain complementary strengths while promoting agreement in judgment.

The important aspect of discriminator consistency is that discriminators that evaluate feature maps from different network architectures can learn from each other. 
Unlike previous methods~\cite{raghu2021vision, stylegan-xl}, which train multiple discriminators on separate projected features without explicitly coupling them, our discriminator consistency loss enforces agreement between discriminators.
This leads to more stable and robust training by preventing any one discriminator from dominating the feedback process. Encouraging discriminators to converge on a consensus about image quality provides the generator with more uniform and coherent feedback, thereby enhancing the overall stability and performance of the GAN.

To empirically demonstrate the benefits of discriminator consistency, we plot the signed real logits in Fig. \ref{sign_graph}. The signed real logits, denoted as $sign(D(\textbf{x}))$, represent the proportion of the training set that receives positive discriminator outputs. This metric serves as a useful heuristic for assessing discriminator overfitting \cite{karras2020training, sauer2021projected,zhang2025improving}. A lower $sign(D(\textbf{x}))$ value, as observed in Config \textbf{D}, indicates reduced overfitting, suggesting that discriminator consistency contributes to a more generalized and stable training process.

In addition, as shown in Table \ref{config_table}, the recall metric, which distinguishes mode-collapse \cite{borji2022pros}, also improves with the application of discriminator consistency. 
By implementing discriminator consistency, HP-GAN not only stabilizes the training process but also improves the quality of the synthesized images.
A more detailed discussion is provided in Sec. \ref{combination}.

\subsection{FakeTwins}
Rather than imposing additional losses on the discriminator to indirectly improve the performance of the generator, we propose a direct approach that leverages pretrained networks within the GAN training.
We introduce FakeTwins, an SSL based mechanism designed to enhance the generative capabilities of the generator.

To effectively utilize pretrained networks, our approach is grounded in the principles of Barlow Twins \cite{barlow-twins}.
The Barlow Twins method promotes feature decorrelation by ensuring that the cross-correlation matrix of the output features approximates the identity matrix. This encourages orthogonality among features, thereby minimizing redundancy and enhancing the diversity of the learned representations.

We hypothesize that obtaining features, normalizing, computing the cross-correlation matrix, and calculating the Barlow Twins loss using a pretrained network would result in lower output loss values for batches containing a diverse set of images. 
This insight motivates the development of FakeTwins, which applies self-supervised learning (SSL) leveraging pretrained networks to enhance the diversity of the generated images.

In our implementation, we leverage pretrained networks with random projections as encoders for self-supervised learning. Notably, these encoders remain fixed during training, and only the linear head network undergoes training.
Representations extracted from two different encoders, subjected to global average pooling, are concatenated to form the final representations that feed into the linear head network.

The FakeTwins loss is specifically applied to perturbed fake images, thereby exclusively targeting the generator.
And noise-related latent augmentation \cite{fakeclr} is utilized for noise perturbation, which alleviates the discontinuity of latent space.
The formulation of the FakeTwins loss is as follows:
\begin{align}
A^\prime &= T_1(G(\mathbf{z})), \quad 
B^\prime = T_2(G(\mathbf{z} + \hat{\boldsymbol{\epsilon}})), \\
V^{A^\prime} &= \mathrm{Concat}(P_{\mathrm{CNN}}(A^\prime), \, P_{\mathrm{ViT}}(A^\prime)), \\
V^{B^\prime} &= \mathrm{Concat}(P_{\mathrm{CNN}}(B^\prime), \, P_{\mathrm{ViT}}(B^\prime)), \\
\mathcal{L}_{\mathcal{FT}} &= 
\mathcal{L}_{\mathcal{BT}}(\phi(V^{A^\prime}), \, \phi(V^{B^\prime})),
\end{align}
where $\hat{\mathbf{\epsilon}}=l_1\cdot|\mathbf{z}|$ and $l_1$ is a coefficient set to $0.1$. Here, $P_{\mathrm{CNN}}$ and $P_{\mathrm{ViT}}$ represent feature projectors composed of pretrained networks. $\phi$ denotes the linear head network.

Fig. \ref{fid_graph} illustrates the reduction in FID as training progresses, alongside the decreasing FakeTwins loss.
Comparing Config \textbf{D} and \textbf{E}, it is evident that FakeTwins promotes faster convergence and improves FID.

\begin{table}[t]
\centering
\caption{Results on benchmark datasets: FFHQ, LSUN-Bedroom and LSUN-Church. FID is reported as the evaluation metric. The best results are in bold.}
\label{main_result_table}
{
\begin{tabular}{lccc}
\hline  
Method               & FFHQ      & Bedroom     & Church    \\ \hline
ADM \cite{dhariwal2021diffusion}             & 2.57      & 1.90        & -         \\
LDM \cite{rombach2022high}                   & 4.98      & 2.95        & 4.02      \\
SAGAN \cite{zhang2019self}                   & 16.21     & 14.06       & 6.15      \\
FastGAN \cite{fastgan}                       & 12.69     & 8.24        & 8.43      \\
GANformer \cite{hudson2021generative}        & 7.42      & 6.15        & 5.47      \\
StyleGAN2 \cite{karras2019style}             & 4.86      & 4.01        & 4.54      \\
StyleGAN3 \cite{karras2021alias}             & 3.80      & -           & -         \\
Dual Contrastive \cite{yu2021dual}           & 4.63      & 3.31        & 3.39      \\
Diffusion GAN \cite{wang2022diffusion}       & 3.73      & 1.43        & 1.85     \\
Projected GAN \cite{sauer2021projected}      & 3.39      & 1.52        & 1.59      \\
Vision-aided GAN \cite{sauer2021projected}   & 3.30      & -           & 1.72      \\
DynamicD      \cite{yang2022improving}       & 3.53      & 4.01        & 3.87      \\ 
CLR-GAN      \cite{sun2024clr}       & 3.37      & -        & 3.43      \\ 
GGDR \cite{lee2022generator}                     & 3.25      & 3.71        & 2.81      \\
GLeaD \cite{bai2023glead}                    & 2.90      & 2.72        & 2.15      \\
HP-GAN                                       & \textbf{1.69} & \textbf{1.19} & \textbf{1.44} \\ \hline
\end{tabular}
}
\end{table}

The FakeTwins loss according to the composition of the images within a batch is shown in Fig. \ref{batch_ft}, averaged over 100 random augmentations.
We compare $\mathcal{L}_{\mathcal{FT}}$ with respect to the quality, diversity, and degree of Gaussian blurring of the images within the batch.
The results indicate that when all images in a batch are identical as shown in the 1st and 3rd rows of the left column, the FakeTwins loss is relatively high, regardless of whether the image is a simple color or a normal human face.
Conversely, when the batch comprises varied images, even if they are simple color images (2nd row of the left column), the FakeTwins loss is lower. 
Additionally, when the batch consists of similar images generated with noise perturbation rather than identical images, the FakeTwins loss is reduced but remains higher than when the images are diverse (4th row).
The right column shows that the less the Gaussian blur in images, the lower the FakeTwins loss. This finding suggests that images with more detail result in a lower loss value.

In summary, the diversity of images within the batch, along with the abundance of features and details within individual images, leads to a decrease in FakeTwins loss, thereby improving the generator's ability to produce diverse and high quality images.

\subsection{Overall Objective Function}
The overall objective function of HP-GAN is formulated as follows:
\begin{align}
\begin{split}
    \mathcal{L}_D^{\prime}&=\mathcal{L}_D+\lambda_{D}^{f}\mathcal{L}_{\mathcal{DC}}(G(\mathbf{z}))+\lambda_{D}^{r}\mathcal{L}_{\mathcal{DC}}(\mathbf{x}),
\end{split}
    \\ \mathcal{L}_G^{\prime}&=\mathcal{L}_G+\lambda_{G}\mathcal{L}_{\mathcal{DC}}(G(\mathbf{z}))+\lambda_{f}\mathcal{L}_{\mathcal{FT}},
\end{align}
where $\lambda_{D}^{f}$, $\lambda_{D}^{r}$, $\lambda_{G}$ and $\lambda_{f}$ are constants that determine the relative weighting of each component.
In $\mathcal{L}'_{D}$, the discriminator consistency loss $\mathcal{L}_\mathcal{DC}$ is applied both to real images $\mathcal{L}_\mathcal{DC}(\mathbf{x})$ and to generated images $\mathcal{L}_\mathcal{DC}(G(\mathbf{z}))$.
In $\mathcal{L}'_{G}$, $\mathcal{L}_\mathcal{DC}(G(\mathbf{z}))$ is applied to generated images, since the generator is updated through the consistency of discriminator outputs on generated samples.
$\mathcal{L}_{\mathcal{FT}}$ is the FakeTwins loss, which is applied through generated images to the generator.

\begin{figure}[tbp]
\centering
    \includegraphics[width=1\linewidth]{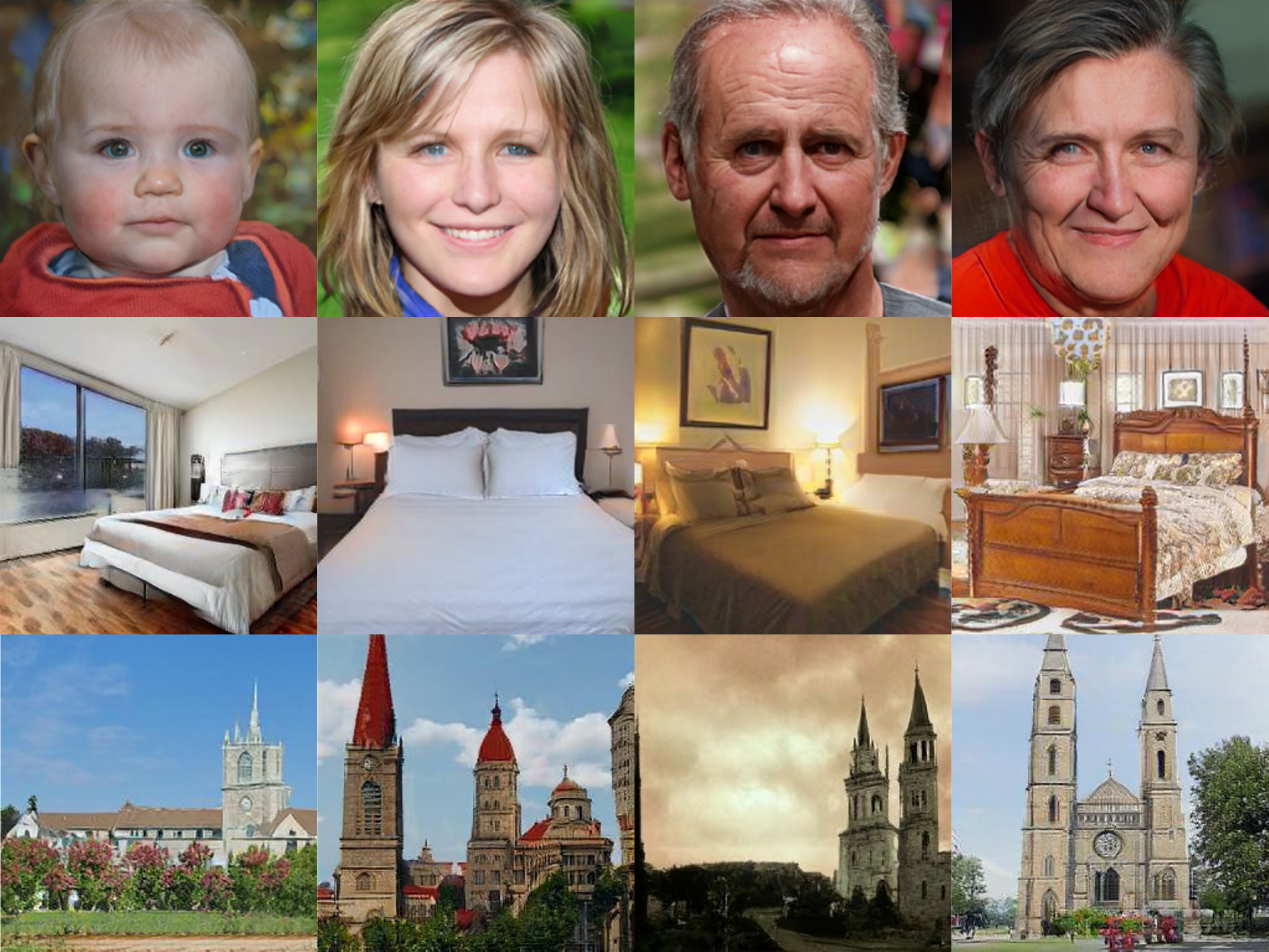}
    \caption{Generated images on benchmark dataset. Categories (top to bottom): FFHQ, LSUN-Bedroom and LSUN-Church.}
    \label{main_result_fig}
\end{figure}

\section{Experiments}
In this section, we evaluate the performance of our proposed HP-GAN by benchmarking it against state-of-the-art models across multiple datasets.
After introducing the implementation details, we present comparisons on various datasets. 
Furthermore, ablation studies investigate the smoothness of the latent space using Perceptual path length (PPL) and the impact of training with limited data by assessing performance on a subset of the FFHQ dataset. Additionally, we explore the effects of different combinations of feature networks by comparing configurations that use only CNNs with those that combine CNN and ViT.

\begin{table}[t]
\centering
\caption{Results on few-shot Datasets. FID is reported as the evaluation metric. The best results are in bold.}
\label{fewshot_table}
\resizebox{\linewidth}{!}{
\begin{tabular}{lccclcc}
\hline
                & \multicolumn{3}{c}{\textbf{100-shot}}      & \multicolumn{1}{c}{} & \multicolumn{2}{c}{\textbf{AnimalFace}} \\ \cline{2-4} \cline{6-7} 
Method         & Obama & \makecell[c]{Grumpy\\Cat} & Panda &  & Cat   & Dog    \\ \hline
EDM + DA \cite{karras2022elucidating}      & 37.10 & 29.94      & 10.81 &  & 36.88 & 57.14 \\
Patch Diffusion \cite{wang2023patch}  & 41.47 & 30.89      & 13.25 &  & 43.71 & 72.17 \\
StyleGAN2 \cite{karras2020analyzing}      & 80.20 & 48.90      & 34.27 &  & 71.71 & 130.2 \\
Scale/shift \cite{noguchi2019image}   & 50.72 & 34.20      & 21.38 &  & 54.83 & 83.04  \\
MineGAN \cite{wang2020minegan}       & 50.63 & 34.54      & 14.84 &  & 54.45 & 93.03  \\
TFGAN \cite{wang2018transferring}   & 48.73 & 34.06      & 23.20 &  & 52.61 & 82.38  \\
TFGAN+DA \cite{zhao2020differentiable} & 39.85 & 29.77      & 17.12 &  & 49.10 & 65.57  \\
FreezeD \cite{mo2020freeze}       & 41.87 & 31.22      & 17.95 &  & 47.70 & 70.46  \\
StyleGAN2-DA \cite{zhao2020differentiable}            & 46.87 & 27.08      & 12.06 &  & 42.44 & 58.85  \\
FastGAN \cite{fastgan}       & 41.05 & 26.65      & 10.03 &  & 35.11 & 50.66  \\
StyleGAN2-ADA \cite{karras2020training}           & 45.69 & 26.62      & 12.90 &  & 40.77 & 56.83  \\
LeCam GAN \cite{tseng2021regularizing}     & 33.16 & 24.93      & 10.16 &  & 34.18 & 54.88  \\
GenCo \cite{cui2022genco}         & 32.21 & 17.79      & 9.49  &  & 30.89 & 49.63  \\
InsGen \cite{insgen}         & 32.42 & 22.01      & 9.85  &  & 33.01 & 44.93  \\
FakeCLR \cite{fakeclr}       & 26.95 & 19.56      & 8.42  &  & 26.34 & 42.02  \\ 
Projected GAN \cite{sauer2021projected}       & 11.21 & 15.80      & 3.98  &  & 18.01 & 17.88  \\ 
Diffusion GAN \cite{wang2022diffusion}       & 10.54 & 15.13      & 3.39  &  & 17.86 & 17.22  \\ 
DANI \cite{zhang2024improving}       & 10.08 & 14.92      & 3.04  &  & 17.72 & 16.81  \\ 
QADDRS \cite{zhang2025improving} & \textbf{9.93} & 14.72      & \textbf{3.08}  &  & 16.94 & 16.57  \\ 
HP-GAN & 10.30 & \textbf{13.21} & 3.34 &     & \textbf{14.34}       & \textbf{15.68}       \\ \hline
\end{tabular}
}
\end{table}

\begin{figure}[tbp]
\centering
    \includegraphics[width=1\linewidth]{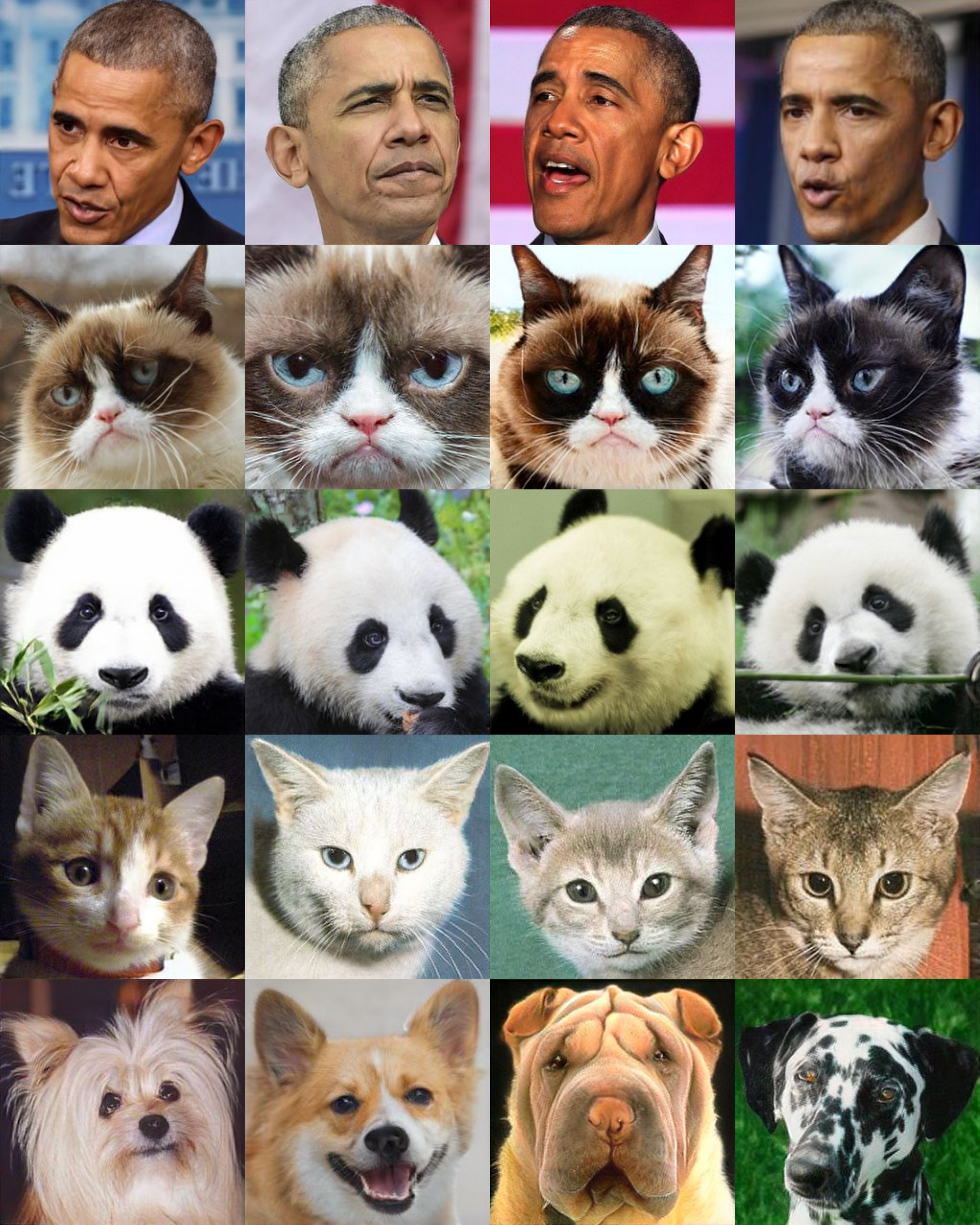}
    \caption{Generated images on few-shot datasets. Top to bottom (Obama, Grumpy Cat, Panda, AnimalFace Cat, AnimalFace Dog)}
    \label{few_fig}
\end{figure}

\subsection{Experimental Setup}
\noindent \textbf{Datasets.} For comparison, all the datasets except AFHQ were resized to $256^2$. We evaluate our approach across a wide range of datasets to ensure comprehensive evaluation.

\noindent • Large datasets: We present the comparisons on large datasets such as  FFHQ (70k images) \cite{karras2019style}, LSUN-Bedroom (3M images) \cite{yu2015lsun}, LSUN-Church (126k images) \cite{yu2015lsun}, CLEVR (70k images) \cite{johnson2017clevr}, and Cityscapes (25k images) \cite{cordts2016cityscapes}, with a primary focus on FFHQ, LSUN-Bedroom, and LSUN-Church as benchmark datasets.

\noindent • Small datasets: To evaluate the performance on small datasets, datasets used for the evaluation include art paintings from WikiArt (1000 images; wikiart.org), Oxford Flowers (1360 images) \cite{nilsback2008automated}, landscape photographs (4319 images; flickr.com), AnimalFace Dog (389 images) \cite{si2011learning}, and Pokemon (833 images; pokemon.com).

\noindent • Few-shot datasets: Our experiments also explore few-shot settings using datasets as Obama (100 images) \cite{zhao2020differentiable}, Grumpy Cat (100 images) \cite{zhao2020differentiable}, Panda (100 images) \cite{zhao2020differentiable}, AnimalFace Cat (160 images) \cite{si2011learning}, and AnimalFace Dog (389 images) \cite{si2011learning}.

\noindent • AFHQ datasets: We evaluate the performance on the AFHQ dataset \cite{choi2020stargan}, which consists of around 5000 images per category for cat, dog, and wildlife, all at a resolution of $512^2$.

\noindent\textbf{Implementation.}
For the generator $G$, we adopted the architecture of FastGAN \cite{fastgan, sauer2021projected}. And for the discriminator $D$, we used the architecture of StyleGAN-XL \cite{stylegan-xl}. We used the combination of EfficientNet-lite0 \cite{tan2019efficientnet} and DeiT-B \cite{touvron2021training} for the feature networks.
The linear head network for FakeTwins had three linear layers, each with 512 output units, and the first two layers were followed by batch normalization and rectified linear units.
Exponential moving average of generator weights and Adam \cite{kingma2014adam} optimizer were adopted.
We employed differentiable data augmentation \cite{zhao2020differentiable} to effectively train GAN and also applied it to generate distorted images for FakeTwins.
For all datasets, we applied data amplification via x-flips.
All models were trained with 20M images to ensure convergence, except for the few-shot datasets, where training was extended to 100M images.
The hyperparameters in HP-GAN are set as $\lambda_{D}^{f}=1$, $\lambda_{D}^{r}=1$, $\lambda_{G}=1$ and $\lambda_{f}=0.02$.
All experiments were conducted on 8 NVIDIA RTX A5000 GPUs with 64 batch size.

\begin{table}[t]
\centering
\caption{Results on AFHQ. FID is reported as the evaluation metric. The best results are in bold.}
\label{afhq_table}
\begin{tabular}{lccc}
\hline
         & \multicolumn{3}{c}{\textbf{AFHQ}}                                              \\ \cline{2-4} 
Method   & Cat                  & \multicolumn{1}{c}{Dog} & \multicolumn{1}{c}{Wild} \\ \hline
DDMI \cite{park2024ddmi}        & 4.27 & 8.54                    & - \\
StyleGAN2 \cite{karras2020analyzing}        & 5.13 & 19.4                    & 3.48 \\
ContraD \cite{contraD}         & 3.82 & 7.16                    & 2.54 \\
StyleGAN2-ADA \cite{karras2020training}              & 3.55 & 7.41                    & 3.05 \\
InsGen \cite{insgen}          & 2.60 & 5.44                    & 1.77 \\
Vision-aided GAN \cite{kumari2022ensembling} & 2.44 & 4.60                    & 2.25 \\
Diffusion GAN \cite{wang2022diffusion}    & 2.40 & 4.83                    & 1.51 \\
Projected GAN \cite{sauer2021projected}   & 2.16 & 4.52                    & 2.17 \\ 
HP-GAN   & \textbf{1.81}     & \textbf{3.63}           &    \textbf{1.18}  \\ \hline  
\end{tabular}
\end{table}
\begin{figure}[tbp]
\centering
    \includegraphics[width=1\linewidth]{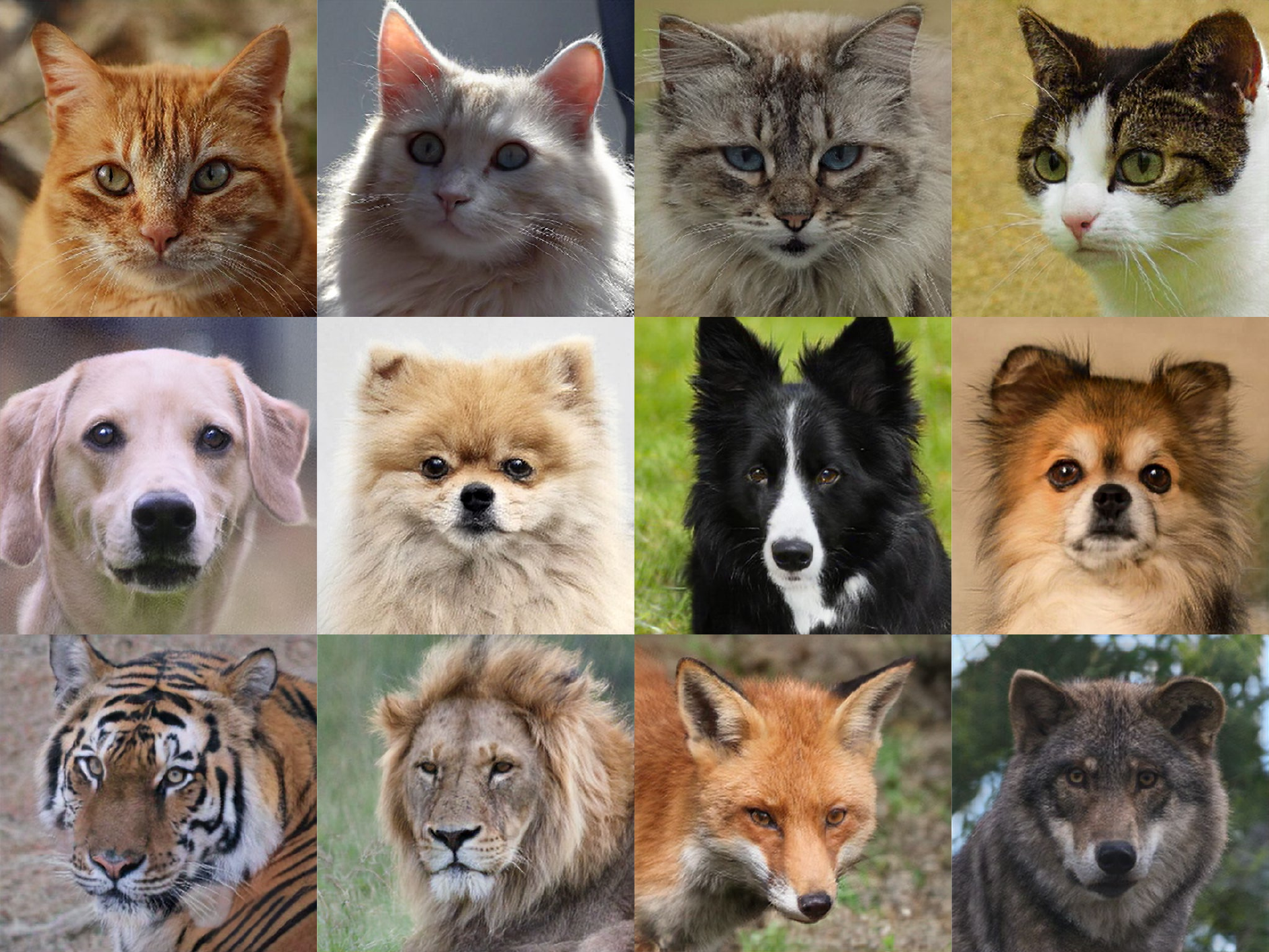}
    \caption{Generated images on AFHQ dataset. Categories (top to bottom): Cat, Dog and Wild.}
    \label{afhq_fig}
\end{figure}

\noindent\textbf{Evaluation.}
We employed the Fréchet Inception Distance (FID) \cite{heusel2017gans} as the main quantitative assessment metric, because FID tends to correlate with human judgement regarding the quality of synthesized images. 
Regardless of the number of training data, we computed the FID between 50k generated images and all training images \cite{karras2019style, karras2020analyzing, fastgan}. 
Our implementation is based on the of official codebase of ADA \cite{karras2020training}. 
And we report the lowest (best) FID, as in the notable studies \cite{sauer2021projected, stylegan-xl}.

Additionally, we compared the precision and recall \cite{kynkaanniemi2019improved} to provide further insights into the fidelity and diversity of the generated images, and Kernel Inception Distance (KID) \cite{binkowski2018demystifying}, which is more suitable for small datasets.

\begin{table}[t]
\centering
\caption{Comparison experiment results for FID, KID, Precsion (Pr), and Recall (Rec) on various datasets. KID indicates the value multiplied by $10^3$ (KID $\times$ $10^3$). Best results are in bold, second best are underlined.}
\label{other_metrics}
\resizebox{\linewidth}{!}{
\begin{tabular}{cccccc}
\hline
Dataset & Method & FID ↓ & KID ↓ & Pr ↑ & Rec ↑ \\ \hline
\multirow{4}{*}{CLEVR} & ADA \cite{karras2020training} & 10.17 & 8.15 & 0.373 & 0.569 \\
 & FastGAN \cite{fastgan} & 3.24 & 2.64 & 0.600 & 0.650 \\
 & Projected GAN \cite{sauer2021projected} & \underline{0.89} & \underline{0.51} & \underline{0.640} & \textbf{0.735} \\
 & HP GAN & \textbf{0.81} & \textbf{0.42} & \textbf{0.693} & \underline{0.718} \\ \hline
 
\multirow{4}{*}{FFHQ} & ADA \cite{karras2020training} & 7.32 & 1.49 & \underline{0.669} & 0.445 \\
 & FastGAN \cite{fastgan} & 12.69 & 5.34 & \textbf{0.716} & 0.184 \\
 & Projected GAN \cite{sauer2021projected} & \underline{3.08} & \underline{0.44} & 0.654 & \underline{0.464} \\
 & HP GAN & \textbf{1.69} & \textbf{0.19} & 0.642 & \textbf{0.545} \\ \hline
 
\multirow{4}{*}{Cityscapes} & ADA \cite{karras2020training} & 8.35 & 3.34 & \textbf{0.649} & 0.146 \\
 & FastGAN \cite{fastgan} & 8.78 & 5.45 & 0.557 & 0.227 \\
 & Projected GAN \cite{sauer2021projected} & \underline{3.41} & \underline{0.91} & \underline{0.619} & \underline{0.361} \\
 & HP GAN & \textbf{2.11} & \textbf{0.72} & 0.590 & \textbf{0.469} \\ \hline
 
\multirow{4}{*}{Bedroom} & ADA \cite{karras2020training} & 11.53 & 7.42 & 0.429 & 0.202 \\
 & FastGAN \cite{fastgan} & 8.24 & 5.90 & 0.602 & 0.189 \\
 & Projected GAN \cite{sauer2021projected} & \underline{1.52} & \textbf{0.36} & \textbf{0.614} & \textbf{0.346} \\
 & HP GAN & \textbf{1.19} & \underline{0.39} & \underline{0.604} & \underline{0.321} \\ \hline
 
\multirow{4}{*}{Church} & ADA \cite{karras2020training} & 5.85 & 4.70 & 0.565 & 0.416 \\
 & FastGAN \cite{fastgan} & 8.43 & 4.61 & \textbf{0.645} & 0.207 \\
 & Projected GAN \cite{sauer2021projected} & \underline{1.59} & \underline{0.50} & \underline{0.612} & \underline{0.438} \\
 & HP GAN & \textbf{1.44} & \textbf{0.48} & 0.577 & \textbf{0.459} \\ \hline
 
\multirow{4}{*}{\begin{tabular}[c]{@{}c@{}}Art\\ Painting\end{tabular}} & ADA \cite{karras2020training} & 43.07 & 10.23 & 0.691 & \underline{0.218} \\
 & FastGAN \cite{fastgan} & 44.02 & 13.00 & \textbf{0.858} & 0.044 \\
 & Projected GAN \cite{sauer2021projected} & \underline{27.96} & \underline{1.25} & 0.762 & \textbf{0.239} \\
 & HP GAN & \textbf{26.31} & \textbf{0.62} & \underline{0.784} & 0.131 \\ \hline
 
\multirow{4}{*}{Landscape} & ADA \cite{karras2020training} & 15.99 & 4.39 & 0.709 & 0.213 \\
 & FastGAN \cite{fastgan} & 16.44 & 3.40 & \underline{0.768} & 0.160 \\
 & Projected GAN \cite{sauer2021projected} & \underline{6.92} & \underline{1.30} & \textbf{0.774} & \underline{0.258} \\
 & HP GAN & \textbf{6.08} & \textbf{0.28} & 0.716 & \textbf{0.425} \\ \hline
 
\multirow{4}{*}{\begin{tabular}[c]{@{}c@{}}AnimalFace\\ Dog\end{tabular}} & ADA \cite{karras2020training} & 60.90 & 22.52 & 0.841 & 0.036 \\
 & FastGAN \cite{fastgan} & 62.11 & 22.11 & 0.849 & 0.015 \\
 & Projected GAN \cite{sauer2021projected} & \underline{17.88} & \textbf{0.03} & \textbf{0.998} & \underline{0.095} \\
 & HP GAN & \textbf{16.38} & \underline{0.25} & \underline{0.993} & \textbf{0.242} \\ \hline
 
\multirow{4}{*}{Flowers} & ADA \cite{karras2020training} & 21.66 & 3.56 & 0.731 & 0.095 \\
 & FastGAN \cite{fastgan} & 26.23 & 6.61 & 0.611 & \underline{0.100} \\
 & Projected GAN \cite{sauer2021projected} & \underline{13.86} & \underline{0.38} & \underline{0.816} & 0.058 \\
 & HP GAN & \textbf{12.03} & \textbf{0.30} & \textbf{0.818} & \textbf{0.107} \\ \hline
 
\multirow{4}{*}{Pokemon} & ADA \cite{karras2020training} & 40.38 & 13.49 & 0.735 & 0.197 \\
 & FastGAN \cite{fastgan} & 81.86 & 80.30 & 0.731 & 0.004 \\
 & Projected GAN \cite{sauer2021projected} & \underline{26.36} & \underline{1.32} & \textbf{0.809} & \underline{0.259} \\
 & HP GAN & \textbf{23.62} & \textbf{1.12} & \underline{0.768} & \textbf{0.310} \\ \hline 
 
\multirow{4}{*}{\begin{tabular}[c]{@{}c@{}}AFHQ\\ Cat\end{tabular}} & ADA \cite{karras2020training} & 3.55 & 0.63 & \underline{0.767} & 0.411 \\
 & FastGAN \cite{fastgan} & 4.69 & 1.72 & \textbf{0.784} & 0.305 \\
 & Projected GAN \cite{sauer2021projected} & \underline{2.16} & \underline{0.16} & 0.693 & \underline{0.565} \\
 & HP GAN & \textbf{1.81} & \textbf{0.08} & 0.741 & \textbf{0.580} \\ \hline
 
\multirow{4}{*}{\begin{tabular}[c]{@{}c@{}}AFHQ\\ Dog\end{tabular}} & ADA \cite{karras2020training} & 7.40 & 1.21 & \textbf{0.753} & 0.470 \\
 & FastGAN \cite{fastgan} & 13.09 & 5.51 & \underline{0.746} & 0.380 \\
 & Projected GAN \cite{sauer2021projected} & \underline{4.52} & \underline{0.80} & 0.718 & \underline{0.643} \\
 & HP GAN & \textbf{3.63} & \textbf{0.37} & 0.737 & \textbf{0.677} \\ \hline
 
\multirow{4}{*}{\begin{tabular}[c]{@{}c@{}}AFHQ\\ Wild\end{tabular}} & ADA \cite{karras2020training} & 3.05 & 0.47 & \textbf{0.765} & 0.137 \\
 & FastGAN \cite{fastgan} & 3.14 & 0.74 & \underline{0.761} & 0.201 \\
 & Projected GAN \cite{sauer2021projected} & \underline{2.17} & \underline{0.12} & 0.705 & \underline{0.292} \\
 & HP GAN & \textbf{1.18} & \textbf{0.02} & 0.737 & \textbf{0.367} \\ \hline
\end{tabular}
}
\end{table}

\begin{figure*}[tbp]
\centering
    \includegraphics[width=1\textwidth]{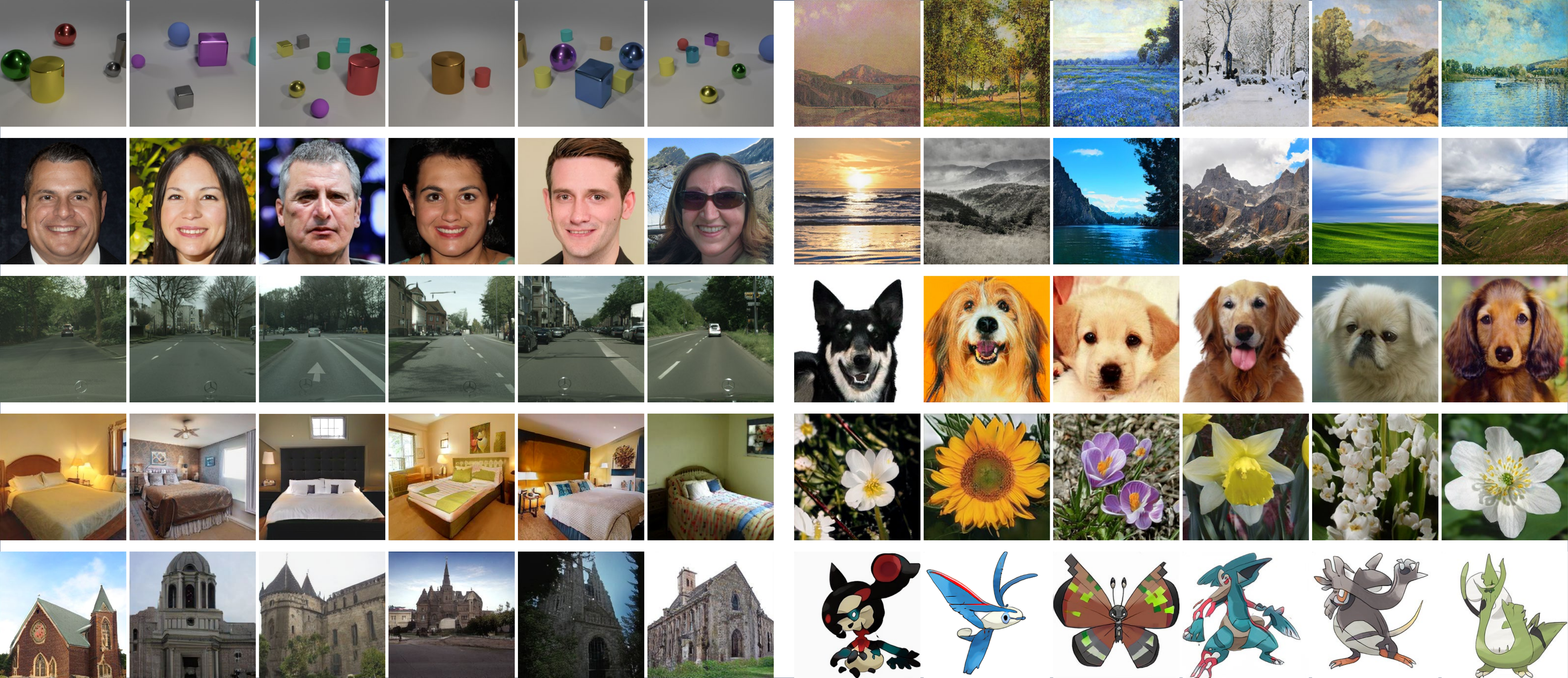}
    \caption{Generated images with various datasets. \textit{Left}: Large datasets (CLEVR, FFHQ, Cityscapes, Bedroom, Church). \textit{Right}: Small datasets (Art Painting, Landscape, AnimalFace Dog, Flowers, Pokemon).}
    \label{results_fig}
\end{figure*}

\subsection{Results}
\noindent\textbf{Benchmark datasets.}
Tab. \ref{main_result_table} presents the comparisons of HP-GAN against various generative models on benchmark datasets. 
HP-GAN consistently achieves superior performance across all datasets.
We compare against both GAN and diffusion methods. 
Among GANs, we include StyleGAN2~\cite{karras2019style}, Projected GAN~\cite{sauer2021projected}, Diffusion GAN~\cite{wang2022diffusion}, GGDR~\cite{lee2022generator}, and GLeaD~\cite{bai2023glead}.
For diffusion models, we evaluate against ADM~\cite{dhariwal2021diffusion} and LDM~\cite{rombach2022high}. 
Notably, HP-GAN achieves the lowest FID scores across all datasets, with a particularly significant improvement on FFHQ, reducing the FID to 1.69.
Qualitative results are shown in Fig.~\ref{main_result_fig}.

\noindent\textbf{Few-shot datasets.}
Few-shot datasets pose significant challenges due to the limited training samples and high risk of overfitting. 
As shown in Tab.~\ref{fewshot_table}, HP-GAN demonstrates consistently strong performance across all datasets, achieving competitive FID scores.
While it does not outperform all methods on every dataset, it achieves state-of-the-art results on AnimalFace Cat and Dog, as well as on Grumpy Cat. 
Qualitative results for few-shot datasets are provided in Fig.~\ref{few_fig}, showing that HP-GAN produces visually consistent and diverse generations even under extreme data constraints.

\noindent\textbf{AFHQ datasets.}
At a resolution of $512^2$, HP-GAN achieves superior performance across all categories of the AFHQ datasets. 
It achieves an FID of 1.81 for the Cat category and 3.63 for the Dog category, outperforming all compared methods. 
Furthermore, HP-GAN attains an FID of 1.18 on the Wild category, highlighting its strong generalization across diverse animal domains.
Quantitative results are presented in Tab. \ref{afhq_table} and generated images are shown in Fig. \ref{afhq_fig}.

\noindent\textbf{Various Datasets.}
To assess the robustness of HP-GAN across diverse conditions and dataset sizes, we compare it against strong baselines, including ADA~\cite{karras2020training}, FastGAN~\cite{fastgan}, and Projected GAN~\cite{sauer2021projected}. 
The results, evaluated using FID, KID, precision (Pr), and recall (Rec), are summarized in Tab.~\ref{other_metrics}. 
For KID, values are scaled by $10^3$, and arrows (↑/↓) indicate whether higher or lower is better. Representative qualitative results are shown in Fig.~\ref{main_result_fig}.

To complement FID, we report Kernel Inception Distance (KID) \cite{binkowski2018demystifying}, which is an unbiased alternative that tends to be more stable, especially on smaller datasets.
Except for two datasets, the KID rankings are consistent with those of FID, indicating that HP-GAN’s performances is not overly tuned to a specific metric.

We further evaluate generation quality and diversity using precision, which measures the fidelity of generated images, and recall, which quantifies the coverage of the real distribution~\cite{kynkaanniemi2019improved}. 
These two metrics reflect the trade-off: low precision suggests degraded image quality, while low recall indicates mode collapse~\cite{borji2022pros}.

Some methods achieve better performance in precision compared to HP-GAN.
Methods like ADA~\cite{karras2020training} and FastGAN~\cite{fastgan}, though strong in precision (fidelity), struggle with recall, suggesting susceptibility to mode collapse. 
In contrast, HP-GAN achieves consistently high recall across most datasets, which explains its strong performance in FID and KID. These results highlight HP-GAN’s ability to maintain high fidelity while preserving sample diversity.

On high-resolution datasets such as AFHQ, both HP-GAN and Projected GAN~\cite{sauer2021projected} show slightly reduced precision. However, HP-GAN surpasses Projected GAN in both precision and recall, demonstrating improvements in image quality and diversity as well as further alleviation of mode collapse.

\section{Ablation Study}
\label{Ablation}
\subsection{Perceptual path length (PPL)}
PPL \cite{karras2019style} is a metric introduced to evaluate the smoothness of the latent space, calculating the average LPIPS distances \cite{zhang2018unreasonable} between generated images when small perturbations is applied to the latent space. This metric also correlates with the consistency and stability of shapes and higher overall image quality \cite{karras2020analyzing}.
As shown in Tab. \ref{config_table}, we compute PPL in latent space, $\mathcal{Z}$, for the entire images without the central crop. Assessments are conducted on configurations where $\mathbf{z} \in \mathbb{R}^{64}$.

Our proposed methods, FakeTwins and discriminator consistency, contribute to enhancing the diversity of the generated images.
And the improvement of PPL demonstrates that the incorporation of these techniques not only enables the generation of diverse images but also ensures high quality.
This indicates that HP-GAN effectively balances  the generation of diverse images while maintaining superior image quality.

\begin{table}[t]
\centering
\caption{Ablation study on FFHQ Subsets. FID is reported as the evaluation metric. The best results are in bold.}
\label{subset_table}
\begin{tabular}{lcccc}
\hline
                & \multicolumn{4}{c}{FFHQ}                                  \\ \cline{2-5} 
Method          & 100          & 1k           & 2k           & 5k           \\ \hline
EDM + DA \cite{karras2022elucidating}          & 50.73          & 30.75        & 27.17        & 24.51        \\
Patch Diffusion \cite{wang2023patch}          & 44.45          & 28.03        & 25.32        & 22.63       \\

StyleGAN2 \cite{karras2020analyzing}      & 179.00          & 100.16       & 54.30         & 49.68        \\
GenCo \cite{cui2022genco}          & 148.00          & 65.31        & 47.32        & 27.96        \\
StyleGAN2-DA \cite{zhao2020differentiable}   & 61.91         & 25.66        & 24.43        & 10.45        \\
StyleGAN2-ADA \cite{karras2020training}   & 85.80         & 21.29        & 15.39        & 10.96        \\
APA \cite{jiang2021deceive}            & 65.00           & 18.89        & 16.90        & 8.38         \\
InsGen \cite{insgen}         & 45.75        & 18.21        & 11.47        & 7.83            \\
FakeCLR \cite{fakeclr}        & 42.56        & 15.92        & 9.90         & 7.25         \\ 
DANI \cite{zhang2024improving}       & 23.98        & 10.81        & 7.73         & 6.20         \\ 
Projected GAN \cite{sauer2021projected}       & 26.25        & 11.12        & 8.25         & 6.85         \\ 
Diffusion GAN \cite{wang2022diffusion}       & 25.47        & 8.76        & 7.99         & 6.59         \\ 
QADDRS \cite{zhang2025improving}       & \textbf{19.89}        & 6.57        & 5.18         & 4.69         \\ 
HP-GAN & 22.20 & \textbf{5.22} & \textbf{3.62} & \textbf{2.86} \\ \hline
\end{tabular}
\end{table}

\subsection{Data efficiency Evaluation}
To evaluate the model's efficiency with limited training data, we conduct experiments using different subset sizes of the FFHQ dataset by randomly sampling with 100, 1k, 2k, and 5k images. Regardless of the number of training data, FID is calculated between 50k generated images and all FFHQ (70k images).
The results are shown in Tab. \ref{subset_table}.

HP-GAN consistently outperforms existing methods across all subset sizes, except for the 100 sample case, where it ranks second.
The improvement in FID is remarkable, demonstrating the data efficiency and effectiveness of the model even when limited data is available
Notably, HP-GAN achieves comparable performance compared to other methods trained on the entire dataset (Tab. \ref{main_result_table}), even when only using 2k or 5k images.
This highlights the robustness and capability of HP-GAN in achieving high-quality image generation with limited data.

\subsection{Feature Networks Combination}
\label{combination}
The combination of feature networks is an important part of HP-GAN. Our ablation studies on FFHQ dataset, detailed in Tab. \ref{fn_table}, compare two distinct combinations: dual CNNs using EfficientNet-lite0 and ResNet50, and the hybrid of CNN \& ViT using EfficientNet-lite0 and DeiT-B. 

The most significant improvement is observed when discriminator consistency is incorporated, as seen in Config \textbf{D} and \textbf{E}.
The enhanced performance achieved with the CNN and ViT hybrid indicates that discriminators are more effectively trained when leveraging feature networks with different architectures. 
In contrast, with dual CNNs, the effect of discriminator consistency is small because both networks process images with similar mechanisms.

Furthermore, SSL using both the ViT and CNN simultaneously shows higher performance than using the same network architecture \cite{song2023multi}.
Additionally, when training networks with distillation, performance is better when the teacher and student networks have different architectures \cite{touvron2021training}.
These findings support the advantage of hybrid feature networks, which benefit from the strengths of both CNN and ViT through the application of discriminator consistency, as discussed in Sec. \ref{D_Consistency}.

\begin{table}[t]
\centering
\caption{Ablation study for feature networks on FFHQ dataset. The combination of CNN \& CNN and CNN \& ViT. FID is reported as the evaluation metric. The best results are in bold.}
\label{fn_table}
\begin{tabular}{lcc}
\hline
\multicolumn{1}{c}{} & \multicolumn{2}{c}{Feature Networks} \\ \cline{2-3} 
Method & CNN \& CNN & CNN \& ViT \\ \hline
StyleGAN-XL \cite{stylegan-xl} & - & 2.19 \\ \hline
Config \textbf{C} & 2.66 & 2.29 \\
Config \textbf{D} & 2.61 & 1.86 \\
Config \textbf{E} & 2.25 & \textbf{1.69} \\ \hline
\end{tabular}
\end{table}

\subsection{Choice of SSL Objective for FakeTwins}
\label{subsec:ssl_objective_ablation}

FakeTwins can be implemented with different SSL objectives.
We compare several choices on FFHQ-1k, keeping all GAN hyper-parameters fixed
and changing only the SSL loss applied to the generator.
The results are reported in Tab.~\ref{tab:ssl_ablation}.

\begin{table}[t]
    \centering
    \caption{Ablation study on FFHQ-1k with different SSL objectives used in FakeTwins. The best results are in bold.}
    \label{tab:ssl_ablation}
    \begin{tabular}{l c}
        \hline
        Method & FID $\downarrow$ \\ \hline
        Config \textbf{D} (no SSL loss)               & 5.65 \\ \hline
        FakeTwins w/ SimCLR \cite{chen2020simple}                    & 5.83 \\
        FakeTwins w/ VICReg \cite{bardes2021vicreg}                   & 5.58 \\
        FakeTwins w/ Barlow Twins \cite{barlow-twins}               & \textbf{5.22} \\ \hline
    \end{tabular}
\end{table}

Both VICReg~\cite{bardes2021vicreg} and Barlow Twins~\cite{barlow-twins} improve over the no-SSL baseline, while SimCLR~\cite{chen2020simple} does not.
In our framework, the SSL loss is applied directly to the generator, and
SimCLR, which explicitly separates samples via negative pairs, may
emphasize instance-level discrimination that is less useful from the generator’s
perspective and is therefore not necessarily well suited.

In contrast, VICReg and Barlow Twins are negative-free, information-maximization
methods that encourage invariance between augmented views while increasing
the information content of the embeddings.
This type of objective is more compatible with FakeTwins, as it promotes informative and diverse feature representations of generated images.
Consequently, we adopt Barlow Twins as the SSL objective for FakeTwins.

\section{Conclusion}
In this study, we introduce HP-GAN, a novel approach to GANs that effectively harnesses the power of pretrained networks and self-supervised learning to enhance image synthesis.
Our method leverages FakeTwins to train generator, thereby improving the diversity and quality of images. 
Furthermore, we introduce discriminator consistency, which aligns the outputs of multiple discriminators with different feature network architectures, promoting more stable and robust training.
Our extensive experiments across a variety of datasets, including large, small, and few-shot settings, demonstrate that HP-GAN consistently outperforms state-of-the-art models.
We show that HP-GAN maintains high performance even with limited training data, emphasizing its efficiency and robustness. 
The ablation studies provide evidence that the hybrid use of CNN and ViT in our feature networks significantly enhances performance, validating the benefits of our approach.
Overall, HP-GAN advances the field of generative modeling by integrating pretrained networks and innovative self-supervised learning strategies, offering a powerful tool for generating high-quality images across different contexts.
This work sets a new precedent for future exploration and development of leveraging pretrained networks.

\section*{Acknowledgments}
This work was supported in part by the National Research Foundation of Korea (NRF) grant funded by the Korea government(MSIT) (RS-2025-02215070, RS-2025-02217919, RS-2025-16070382); in part by the MSIT(Ministry of Science, ICT), Korea, under the Global Research Support Program in the Digital Field program (RS-2024-00436680) supervised by the IITP (Institute for Information \& Communications Technology Planning \& Evaluation); in part by Artificial Intelligence Graduate School Program at Yonsei University (RS-2020-II201361); in part by the KIST Institutional Program (2E33801, 2E33800); in part by the Yonsei Signature Research Cluster Program of 2024 (2024-22-0161).

\section*{Appendix A. Fine-tuning from Natural Images to Medical Images.}
Generative models for medical imaging, including GAN- and diffusion-based approaches, have attracted considerable attention for data augmentation, disease diagnosis, anatomical segmentation, and anomaly detection~\cite{sharma2019missing, shin2025physics, shin2025anatomical}.

We investigate whether HP-GAN can be transferred from natural-image
pretraining to medical imagery and whether domain mismatch in the pretrained
feature networks is a problem.
To this end, we conduct an experiment on the BraTS2021~\cite{baid2021rsna}
dataset using only the T1 contrast.
We extract 1{,}151 training slices and compare two settings:
(i) training HP-GAN from scratch on BraTS, and
(ii) fine-tuning HP-GAN initialized from our FFHQ-pretrained model
(natural face images).

\begin{figure}[tbp]
    \centering
    \includegraphics[width=\linewidth]{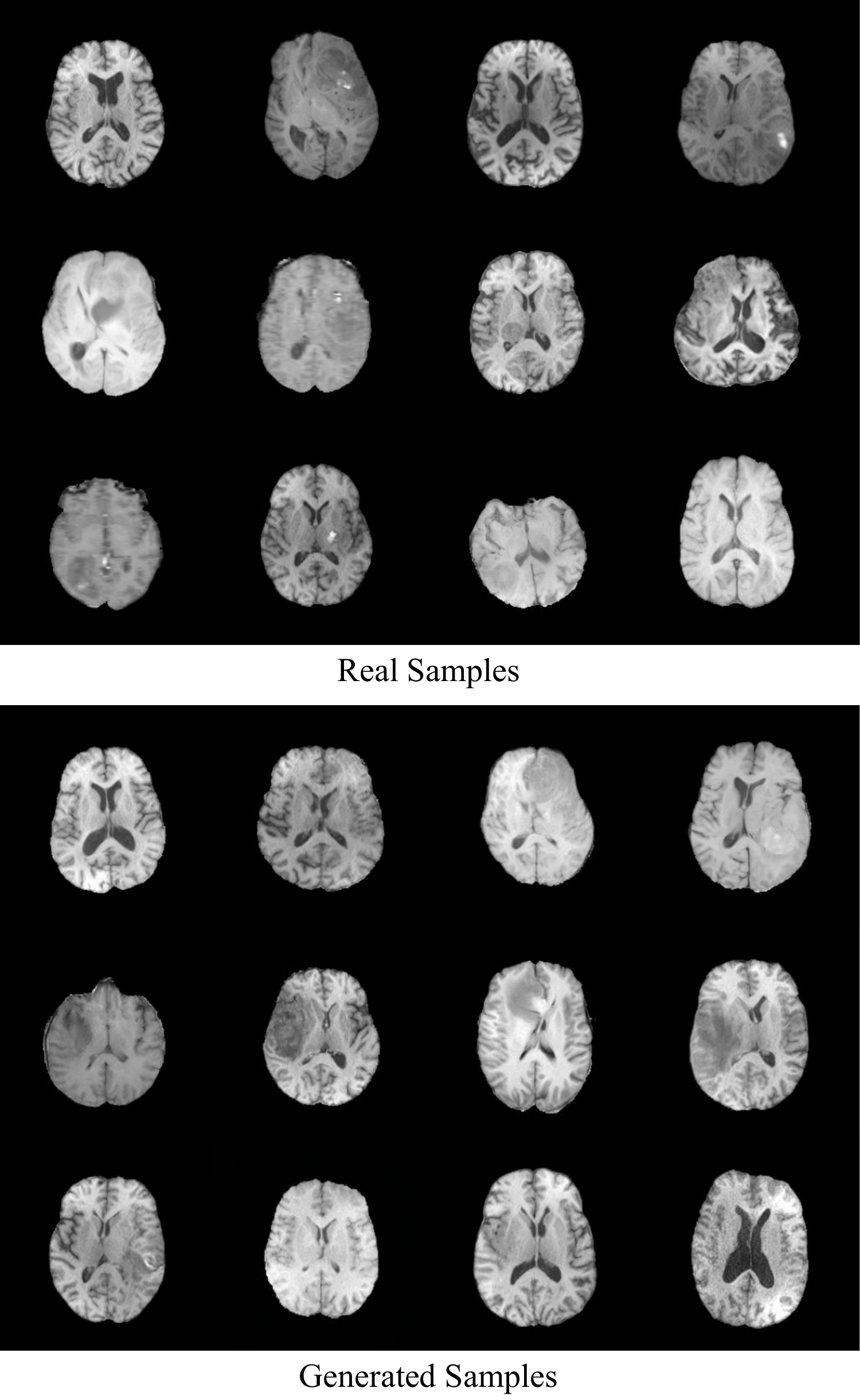}
    \caption{Qualitative results on BraTS2021 T1 contrast. Top: real training samples.
Bottom: images generated by HP-GAN fine-tuned from FFHQ, demonstrating successful transfer from natural to medical images.}
    \label{fig:brats_qual}
\end{figure}

As shown in Tab.~\ref{tab:brats_finetune}, fine-tuning from FFHQ leads to a improvement in FID compared to training from scratch, indicating that HP-GAN benefits from natural-image pretraining even when the target domain is medical imagery.
Fig.~\ref{fig:brats_qual} provides qualitative examples for the fine-tuned model.

\begin{table}[!h]
    \centering
    \caption{Quantitative results on BraTS2021 T1 for HP-GAN trained
    from scratch and fine-tuned from an FFHQ-pretrained model.}
    \label{tab:brats_finetune}
    \begin{tabular}{l c}
        \hline
        HP-GAN setting            & FID $\downarrow$ \\ \hline
        Trained from scratch      & 27.73 \\
        Fine-tuned from FFHQ      & \textbf{9.39} \\ \hline
    \end{tabular}
\end{table}

\section*{Appendix B. Alternative Feature Networks}
\label{app:features}

In the main experiments, we adopt EfficientNet-lite0 and DeiT-B, which were trained for classification, as the default feature networks following StyleGAN-XL~\cite{stylegan-xl}.
To examine the dependence of HP-GAN on the specific choice of feature networks, we evaluate several alternative combinations on FFHQ, varying the pretraining strategy (supervised classification, self-supervised DINO~\cite{caron2021emerging}, and CLIP~\cite{radford2021learning}).
The configurations and FID scores are summarized in Tab.~\ref{tab:feature_networks}.

\begin{table}[t]
    \centering
    \caption{Quantitative results on FFHQ for HP-GAN with different
    combinations of feature networks and pretraining objectives.
    The feature networks are trained with supervised classification (CLS),
    self-supervised DINO, and CLIP pretraining.}
    \label{tab:feature_networks}
    \begin{tabular}{l l c}
        \hline
        CNN         & ViT         & FID $\downarrow$ \\ \hline
        ResNet-50 (CLIP)     & ViT-B (DINO)               & 2.03 \\
        ResNet-50 (CLIP)     & DeiT-B (CLS)               & 1.91 \\
        EfficientNet-lite0 (CLS) & ViT-B (DINO)          & 1.88 \\ 
        EfficientNet-lite0 (CLS) & DeiT-B (CLS)          & \textbf{1.69} \\  \hline
    \end{tabular}
\end{table}

All these configurations yield strong performance, with relatively small differences in FID.
This suggests that HP-GAN is not tied to a particular feature-network
architecture or pretraining objective: both supervised and self-supervised
ViT backbones (e.g., DINO) and CLIP-pretrained CNNs work well, as long as
they provide sufficiently informative features.
This observation is consistent with prior findings in Projected GAN~\cite{sauer2021projected} and
StyleGAN-XL~\cite{stylegan-xl}, where pretrained feature networks perform well regardless of training data, pretraining objective, or network architecture.




\bibliographystyle{cas-model2-names}

\bibliography{cas-refs}

\end{document}